# A salt and pepper noise image denoising method based on the generative classification

Bo Fu[1], Xiao-Yang Zhao[1], Yong-Gong Ren[1], Xi-Ming Li[2], Xiang-Hai Wang[1]
1. College of Computer and Information Technology, Liaoning Normal University, Dalian, China
2. College of Computer Science and Technology, Jilin University, Changchun, China
E-mail: fubo@lnnu.edu.cn

*Abstract*— **In this paper, an image denoising algorithm is proposed for salt and pepper noise. First, a generative model is built on a patch as a basic unit and then the algorithm locates the image noise within that patch in order to better describe the patch and obtain better subsequent clustering. Second, the algorithm classifies patches using a generative clustering method, thus providing additional similarity information for noise repair and suppressing the interference of noise, abandoning those categories that consist of a smaller number of patches. Finally, the algorithm builds a non-local switching filter to remove the salt and pepper noise. Simulation results show that the proposed algorithm effectively denoises salt and pepper noise of various densities. It obtains a better visual quality and higher peak signal-to-noise ratio score than several state-of-the-art algorithms. In short, our algorithm uses a noisy patch as the basic unit, a patch clustering method to optimize the repair data set as well as obtain a better denoising effect, and provides a guideline for future denoising and repair methods.**

*Index Terms—image denoising, patch clustering, salt and pepper noise, non-local switching filter*

## I. INTRODUCTION

Images are always polluted by noise during image acquisition and transmission, resulting in low-quality images. Thus, it is necessary to remove the noise before using the images for subsequent analysis tasks. A very common type of image noise is the so-called salt and pepper noise, which is scattered throughout the image and consists of only the maximum or minimum intensity values (i.e., 0 or 255) in the dynamic range. Generally, the removal of salt and pepper noise consists of two problems: (1) how to detect the noisy pixels and (2) how to repair them. In the past decade, a number of denoising algorithms have been developed. One well-known algorithm family is nonlinear filters such as median filters [1–2]. This family is empirically effective for low density noise. However, it filters all pixels in the image, resulting in the corruption of noise-free pixels. To avoid secondary corruption, the switching filter and its extensions [3–10] introduce a preprocessing step to detect noisy pixels. Boundary discrimination noise detection (BDBD) filter [11] divides images into low-amplitude noisy pixels, high-amplitude noisy pixels, and noise-free pixels. Algorithms based on fuzzy sets [12–13] avoid the false restoration of noise-free pixels by setting fuzzy threshold values.

Recently, a number of algorithms have also attempted to restore noisy pixels using information from the global image and got better denoising performances. Some representative algorithms include non-local means (NLM), block-matching 3D filtering (BM3D), and their extensions [14–18]. However, they cannot be directly applied to denoise salt and pepper noise. Nasri et al. improved classical NLM by introducing a switching filter to reduce the interference of impulse noise [19]. However, because the template is a fixed size, it could not obtains enough useful data in high density noise environments and leads to poor result. Zhang et al. added a decision step to NLM [20] to describe noise, but lacked a basic unit of measure (e.g., patches).

It can be seen that the current prevailing denoising methods mainly focus on improving the following aspects: the noise identification method, noise point repair model, and expansion of the repair information. Although these methods have improved denoising results, the analysis of the above algorithms is heavily dependent on local repair information, i.e., there should be enough information left in the denoised image. When images are corrupted by high density noise, the denoising methods mentioned above usually obtain poor denoise results because they lack enough analysis information. At the same time, simply expanding the scope of repair information cannot obtain an obvious improvement in a high density noise environment. In view of the above problems, the motivation of our research is to optimize the repair information analysis by expanding the search scope, classification patches, and construction of switching templates. The improved denoising method should be able to restrain the interference of noise in the learning process of the repair model and obtain the corresponding useful information for the different content to be repaired.

Given the above motives, a salt and pepper noise image denoising method based on the generative classification is proposed. Our work includes the following two novel steps: First, we define the local patch model and design a generative classification method based on it. Second, using the patches classification, we design a new non-local switching filter. These operations mine data correlation effectively using dimensional reduction. Classification enables the target block to find potential corresponding areas with optimal learned data, which is helpful for model regression. Meanwhile, some clusters that consist of a few patches are abandoned to suppress noise[21-24]. The results of a series of experiments using different cluster numbers, patch sizes, and noise densities show that our method has a better visual effect and higher image performance indicators than classical non-local methods and switching filter methods.

## II. PROPOSED METHOD

The basic framework of the proposed algorithm is shown in Fig. 1. A corrupted image of size $M \times N$ is denoted by **V**, and the corresponding original image is denoted by **U**. Our algorithm estimates **U** using **V**. Its key steps can be summarized as follows.

Step 1: Use a local noise identifier to mark the pixels corrupted by the salt and pepper noise and do not deal with the Gaussian noise.

Step 2: Divide **V** into a series of overlapping patches. The patch size is $L \times L$ and is determined according to the noise density. Then, use the expectation-maximization (EM) method discussed in Section II-A to cluster all the patches and assign each patch a corresponding class label.

Step 3: For each corrupted pixel, perform the non-local switching filtering in the corresponding class.

Step 4: Fuse the denoised pixels and normal pixels into the resulting image.

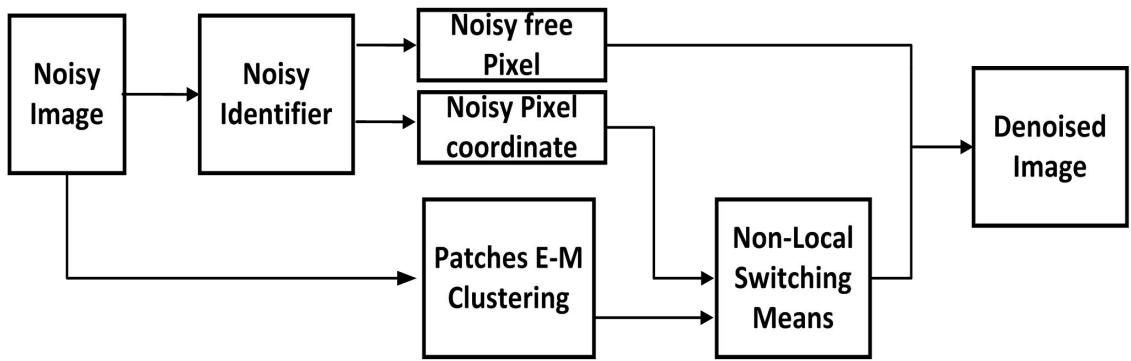

**Fig. 1. Algorithm flow chart**

Let $(i, j)$ be the coordinates of a pixel. The noise model is then defined as:

$$f_{(i,j)} = \begin{cases} N_p & \text{if } f \in [0...T_p) \\ N_s & \text{if } t \in (T_s, 255] \\ V_{(i,j)} & \text{if } t \in (T_p, T_s) \end{cases} \qquad (1)$$

where $N_s$ and $N_p$ are the salt noise and pepper noise, respectively, and $T_s$ and $T_p$ denote the initial thresholds for discriminating the two kinds of noises (roughly determined by a histogram). Of course, if the noise pixels are judged only by thresholds, black or white edges and pure black or white patches will be misjudged. Hence, it is necessary to refine the initial judgments further. So, noise identifier such as Weber's law noise identifier [25] could be added.

*A. Patch modeling and clustering*

We employ a Bayesian model to describe the image patch generation process. Using this model, patches can be classified by the cluster method to find similar patches. Suppose that each image is represented by a mixture of topics, i.e., patches, and each patch follows a Gaussian distribution. To generate an image, a multinomial distribution $\phi$ is first drawn over $K$ topics with Dirichlet prior $\beta$. Then, a topic $z$ is repeatedly chosen from distribution $\phi$ and a patch is generated from the selected topic $\phi_z$ until all patches are generated. This generative process is outlined as follows:

Step 1. Choose $\phi \sim Dirichlet(\beta)$

Step 2. For each of the *n* patches *P*,
    a) Choose a class center $z_k \sim Multinomial(\phi)$.
    b) Sample a patch: $p_n \sim Gauss(\mu_{Z_k}, \Sigma_{Z_k})$

Here, each topic *k* follows a Gaussian distribution with mean $\mu$ and covariance matrix $\Sigma_k$, which is defined as:

$$Gauss(p \mid \mu_k, \Sigma_k) = \frac{1}{(2\pi)^{d/2} |\Sigma_k|^{1/2}} e^{-\frac{1}{2}(p-\mu_k)^T (\Sigma_k)^{-1}(p-\mu_k)} \qquad (2)$$

Our goal is to compute all patches that fall under a similar topic. To increase the sample numbers, we partition the patches point-wise and use EM to infer the parameters of interest: $z, \phi, \{\mu_k\}_{k=1}^{k=K}$, and $\{\Sigma_k\}_{k=1}^{k=K}$.

Using the above patch model, we propose a generation of EM clustering algorithms. In the E-step, fix $\phi$, $\mu$, and $\sigma$, and then compute the expectations for z as follows:

$$h_{nk} := p(z_n = k \mid p_n, \phi, \mu, \Sigma) = \frac{\phi_k Gauss(p_n, \mu_k, \Sigma_k)}{\sum_{i=1}^{K} \phi_i Gauss(p_n, \mu_i, \Sigma_i)} \qquad (3)$$

In the M-step, maximize the expectation of *z* using the following iterative estimation class:

$$\phi_k \leftarrow \left| \sum_{n=1}^{N} w_{nk} + \beta \right| / (N + K\beta) \qquad (4)$$

$$\mu_k \leftarrow \frac{1}{\sum_{n=1}^{N} h_{nk}} \sum_{n=1}^{N} h_{nk} p_n \qquad (5)$$

$$\Sigma_k^v \leftarrow \frac{1}{\sum_{n=1}^{N} h_{nk}} \sum_{n=1}^{N} w_{nk} (p_n - \mu_k)(p_n - \mu_k)^T \qquad (6)$$

Using this EM method, all patches can be categorized into classes, with each patch having a class membership probability of *m*. We can thus compute the most similar patches over the whole image.

We believe the proposed model is of much benefit to denoising. On one hand, the patch model describes an image at a small local scale and can be approximated by a simple Gaussian distribution. On the other hand, each small patch still retains a certain structure and the content features of the whole image. Obviously, the similarity between structures is more robust than the similarity between pixels. In addition, salt and pepper noise in patch could be regarded as an outlier of the clustering data. Therefore, the EM clustering method could suppress noise potentially, so that more similar patch can be searched and data regression training can be better carried out.

*B. Non-local switching filter based EM clustering*

All corrupted pixels are determined and the patches are classified in the procedures described in Section A. The corrupted points are restored using two main steps, namely, candidate set generation and non-local switching filtering.

Specifically, for a corrupted pixel $I_{(i,j)}$, we find the target patch **P** with $I_{(i,j)}$ as the center and the class labels $l$ of **P**. Furthermore, we use the label of **P** to search for similar patches with the same label. If the number of similar patches is insufficient, patches with other similar labels are automatically searched for. Thus, a patch set **S** is formed that consists of patches similar to **P**, with each similar patch called a "reference patch".

The classical NLM method is usually degraded by noise during the training process. Therefore, an idea of switching filter [7, 8] is adopted to reduce the effect of noise on the block repair model. Because noise's locations have been marked, we can replace suspected noise with the patch's mean. Further, all patches in set **S** are used to restore the target patch **P** using a switching non-local filter. Let the restored gray value of the corrupted pixel be $\tilde{I}_{(i,j)}$, which is obtained by weighted averaging as follows:

$$\tilde{I}_{(i,j)} = \sum_{\forall (k,l) \in \Omega_n} w_{((i,j),(k,l))} \cdot I_{(i,j)} \qquad (7)$$

where $w_{((i,j),(k,l))}$ is the weight of patch $P_{(k,l)}$ for $P_{(i,j)}$, $P_{(i,j)}$ indicates a patch with pixel $I_{(i,j)}$ as the center, and $0 < w_{((i,j),(k,l))} < 1$. Weight $w_{((i,j),(k,l))}$ is calculated using the similarity $s_{((i,j),(k,l))}$ between each reference patch and the target patch **P** as follows:

$$w_{((i,j),(k,l))} = \frac{s_{((i,j),(k,l))}}{\sum_{1}^{n} s_{((i,j),(k,l))}} \qquad (8)$$

$$s_{((i,j),(k,l))} = \frac{1}{e^{\frac{\|P_{(i,j)} - P_{(k,l)}\|^2}{\sigma_n}} \ln \sigma_n} \qquad (9)$$

### III. EXPERIMENTS

In this section, we evaluate the performance of our algorithm on salt and pepper noise of various levels of intensity, which describes the degree of corruption. In addition, the denoising results are compared with several main stream denoising methods: adaptive median filter (AMF), decision based algorithm (DBA), modified decision based unsymmetric trimmed median filter (MDBUTMF), switching non-local means (SNLM), and boundary discrimination noise detection (BDND). For fair comparison, all methods' code implementations are their publicly available versions and their parameters are set following the guidelines in original articles. The peak signal-to-noise ratio (PSNR) is adopted to measure the objective performance of our algorithm. Experiments were carried out on corrupted images with various noise densities (10%–90%). The denosing results of Image Lenna are compared in Table 1. Where it can be seen that our algorithm obtains best score at 10% to 50%, and 90% noise levels.

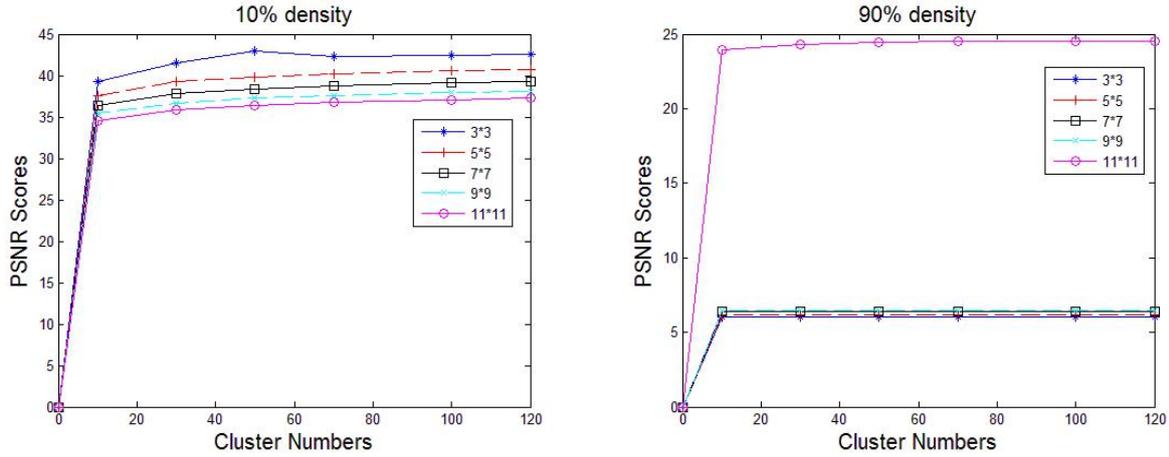

Fig. 2　Results for different size of templates

Furthermore, to investigate the influence of cluster number and template size on the denoising performance, we tested different template sizes (3 to 11) and cluster numbers (10 to 120). The results are shown in Fig. 2. In terms of template size, we observe that our method performs better for lower noise density when the size is small (i.e., 3), whereas better for higher noise density when the size is large (i.e., 11). That is because small templates contain enough useful information for lower noise, but not for higher noise. In higher noisy context (e.g., 90%), many small templates even contain no noise-free pixels, thus we prefer larger template. In terms of cluster number, our method prefers cluster larger number. In the case of cluster number greater than 100, PSNR score grows slowly. Considering the time complexity, 100-500 classes is worthwhile to experiment.

Table 1 PSNR scores result

| level | AMF | DBA | MDBU-TMF | SNLM | BDND | Our Method's PSNR/SSIM |
|---|---|---|---|---|---|---|
| 10% | 38.61 | 36.40 | 37.91 | 39.25 | 42.49 | **42.77/ 0.9977 (3,300)** |
| 20% | 36.10 | 32.90 | 34.78 | 36.65 | 38.48 | **39.33 / 0.9947 (3,300)** |
| 30% | 33.79 | 30.15 | 32.29 | 35.13 | 35.96 | **36.44/ 0.9895 (4,300)** |
| 40% | 32.22 | 28.49 | 30.32 | 33.66 | 34.23 | **34.66/ 0.9831 (4,300)** |
| 50% | 30.39 | 26.41 | 28.18 | 31.96 | 32.52 | **32.59/ 0.9703 (5,300)** |
| 60% | 28.81 | 24.83 | 26.43 | 29.93 | **30.82** | 30.26/ 0.9482 (6,150) |
| 70% | 27.11 | 22.64 | 24.30 | 27.75 | **29.56** | 28.68/ 0.9191 (7,150) |
| 80% | 24.89 | 20.32 | 21.70 | 25.61 | **27.68** | 27.01/ 0.8741 (8,150) |
| 90% | * | 17.14 | 18.40 | 23.47 | * | **24.53/ 0.7816 (11,150)** |

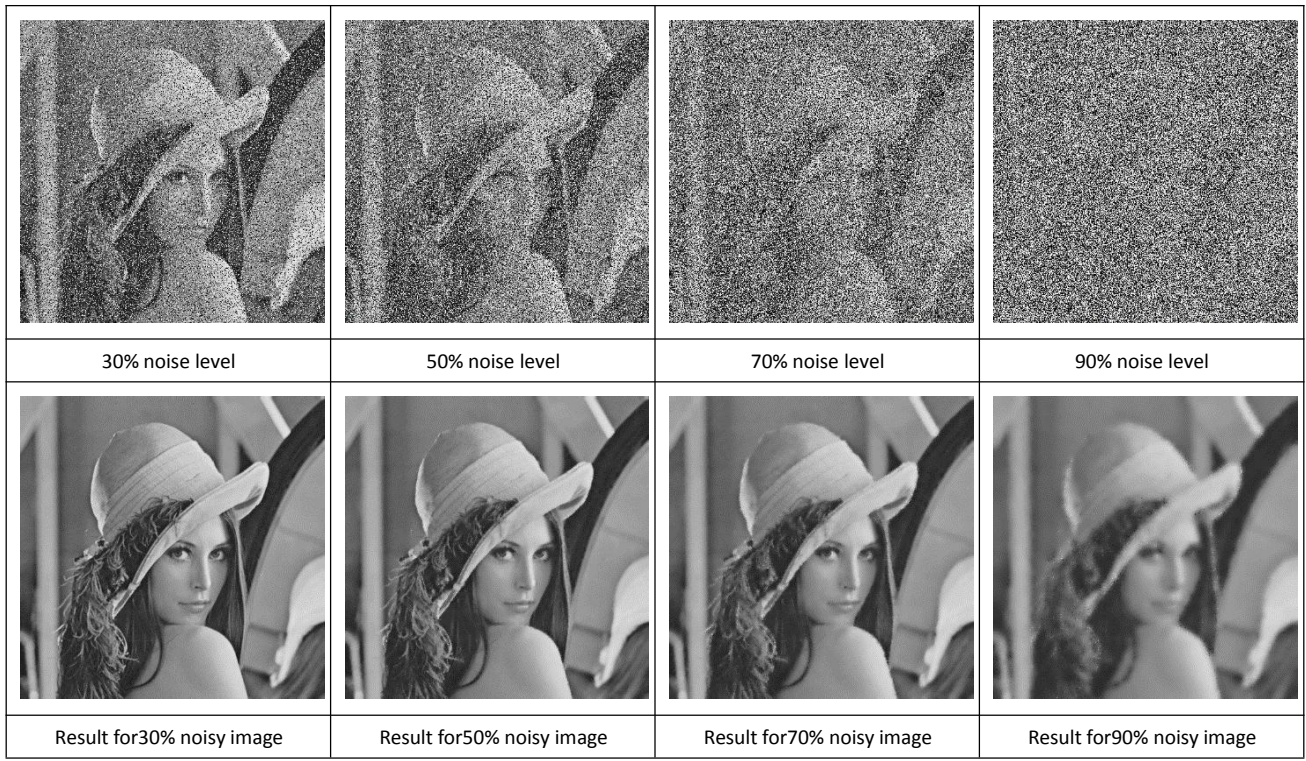

<div style="text-align:center">Fig. 3 Visual quality results of Image Lenna</div>

We also tested images with different textures, include Image Square, Bacteria, Barb, Baboon, Bridge and Bark. The visual effects of denoising results are shown in Fig4, all images are polluted by 30% level noise. The PSNR and SSIM scores of this algorithm for different images and different intensity noise are given in Table 2. It can be seen that good denoising effects are obtained. At the same time, we find that there is a certain correlation between the image noise intensity and the texture content when the size of the patch and the number of classes are selected. Smooth texture image tend to prefer larger size patch that is because less variety of the contents produced by a smooth texture image. High noise level image tend to prefer larger size patch because more noise exist in patch, so more regular pixels are needed.

<div style="text-align:center">Table 2 PSNR scores result for different textures</div>

| Name | 10% PSNR/SSIM | 30% PSNR/SSIM | 50% PSNR/SSIM | 70% PSNR/SSIM | 90% PSNR/SSIM |
|---|---|---|---|---|---|
| Squares | 52.06/0.999 (5,130) | 48.51/0.998 （5,110） | 36.98/0.980 (5,110) | 32.87/0.957 (5,110) | 29.79/0.938 (11,110) |
| Bacteria | 38.06/0.994 (3,110) | 33.97/0.977 (4,110) | 30.73/0.958 (5,110) | 28.73/0.964 (7,110) | 26.41/0.923 (11,90) |
| Barb | 35.64/0.980 (5,110) | 30.14/0.932 (5,110) | 27.04/0.868 (5,110) | 24.45/0.758 (7,110) | 21.72/0.582 (11,110) |
| bridge | 31.73/0.969 (5,110) | 25.55/0.860 (9,110) | 24.13/0.850 (9,110) | 21.98/0.587 (11,110) | 20.40/0.414 (11,110) |
| Baboon | 31.21/0.969 (3,110) | 26.25/0.896 (3,110) | 23.18/0.767 (5,110) | 20.84/0.576 (9,110) | 19.24/0.367 (11,110) |
| Bark | 27.98/0.983 (3,110) | 25.93/0.936 (3,110) | 22.85/0.816 (5,110) | 19.45/0.573 (9,110) | 17.31/0.322 (11,110) |

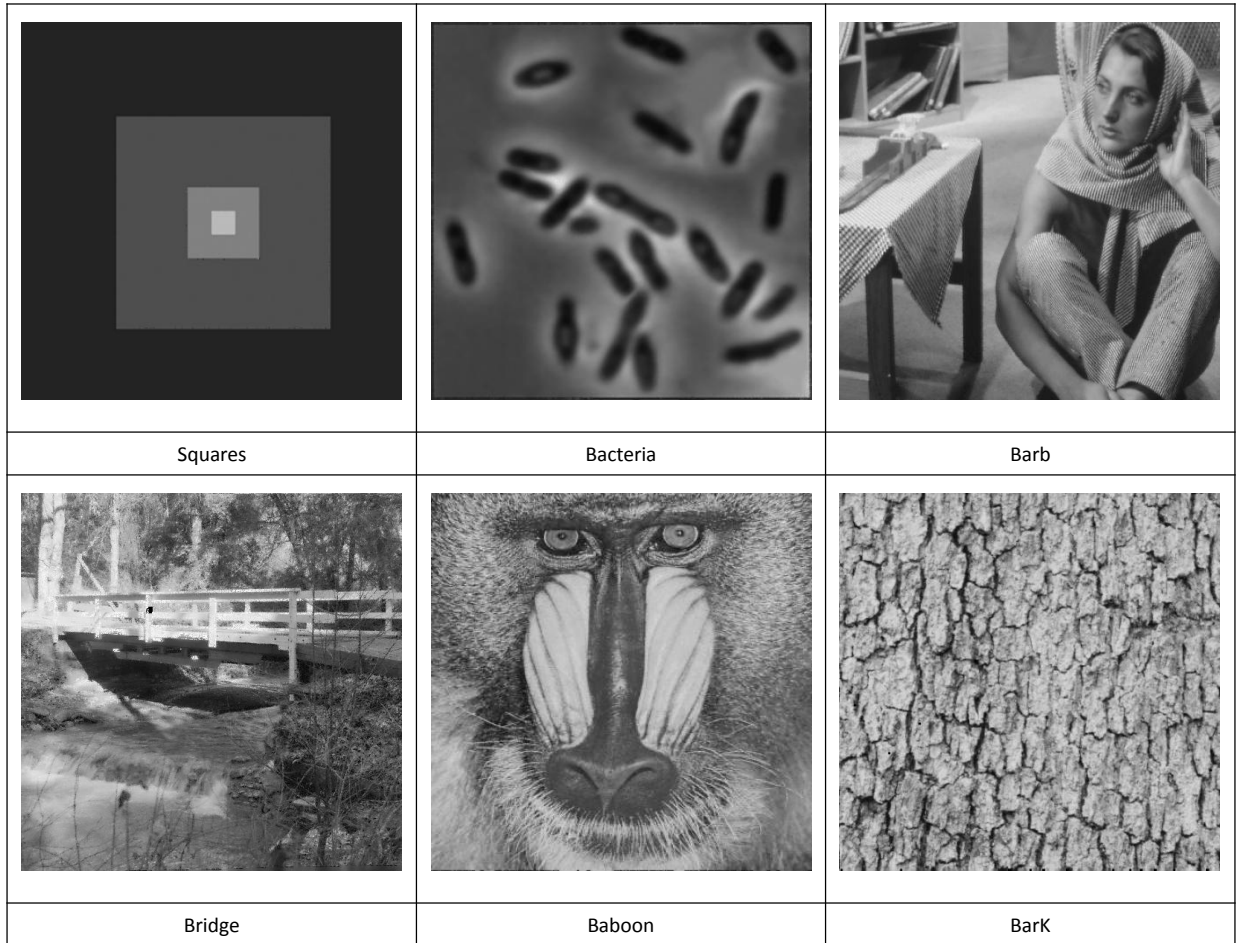

**Fig. 4 Visual quality results of different textures images**

## IV. Conclusion

In this paper, an image denoising algorithm based on generative classification was proposed for salt and pepper noise removal. First, a generative model is built on a patch as a basic unit and then the algorithm locates the positioning image noise within that patch, in order to facilitate a better description of the patch and the subsequent clustering. Second, patches are classified using a generative clustering method, thus providing additional similarity repair information for noise repair and suppressing noise interference by abandoning small categories. Finally, the algorithm builds a non-local switching filter to remove the salt and pepper noise. Simulation results show that the proposed algorithm is an effective way to denoise salt and pepper noise with various densities. It obtains a better visual quality and a higher PSNR score at some noise density levels than state-of-the-art algorithms.

## Acknowledgements

This work is supported by the National NaturalScience Foundation of China (NSFC) Grant No. 61702246, 61402214, 61602204, and 41671439, Liaoning Province of China General Project of Scientific Research No. L2015285, Liaoning Province of China Doctoral Research Fund No. 201601243, and Liaoning University Youth Project No.LS2014L014.